\documentclass{article}
\usepackage{spconf,amsmath,graphicx,hyperref}

\usepackage{cleveref}

\newcommand{\RomanNum}[1]{\uppercase\expandafter{\romannumeral #1}}
\usepackage{amssymb}
\usepackage[table]{xcolor}
\usepackage{booktabs}
\usepackage{graphicx} 
\usepackage{multirow} 
\usepackage{subcaption}
\usepackage[ruled,vlined]{algorithm2e}
\usepackage{amsmath} 
\title{Adversarial Consistency-Guided Representation Learning for Multi-View Clustering}
\name{
    Yuchen Lin  $^{1,2,\dagger}$ \qquad
    Kunpeng Xu $^{3,\dagger} $ \qquad
    Ying Fang $^{1,*}$ \qquad
    Lifei Chen $^{1,2, *}$ \qquad
    \thanks{$^\dagger$ Equal contribution. $^*$ Corresponding authors.}
}

\address{
    $^{1}$ School of Computer and Cyber Security, Fujian Normal University, Fuzhou 350000, China \\
    $^{2}$ Digital Fujian Internet-of-Things Laboratory of Environmental Monitoring, \\ Fujian Normal University, Fuzhou 350000, China\\
    $^{3}$ School of Computer Science, McGill University, Montreal H3A 2A7, Quebec, Canada \\
}
\begin{document}
\ninept
\maketitle
\begin{abstract}
Multi-view clustering aims to capture cross-view consistency while exploiting view-specific information. However, shared representations learned to capture cross-view consistency may still retain view-identifying information, potentially compromising the consistency of cross-view clustering structures. To address this issue, we propose ACGRL, an adversarial consistency-guided representation learning framework for multi-view clustering. ACGRL employs a gradient-reversal view discriminator to reduce view identifiability and obtain invariant reference representations. These representations are then frozen to provide fixed references for disentangling view-specific information from cross-view common information in the subsequent learning stage. The fixed reference representations are concatenated with the learned view-specific representations for reconstruction and clustering, with cross-view cluster alignment encouraging consistent clustering assignments. Experiments on four benchmark datasets demonstrate the superior clustering performance of ACGRL compared with representative multi-view clustering methods.
\end{abstract}
\section{Introduction}
\label{Section_Introduction}

Multi-view clustering aims to discover cluster structure by exploiting
cross-view consistency and complementary information across
heterogeneous views \cite{Self_ExpressiveMetrixLearning}. Existing methods pursue this goal through representation alignment, dependence maximization, or shared-specific decomposition \cite{DMVCS,DFL_Net}. However, shared representations may still retain view-identifying information, potentially compromising the consistency of cross-view clustering structures. To address this limitation, we propose ACGRL, an adversarial consistency-guided representation learning framework that uses fixed references to guide the separation of common and view-specific information. In Stage~\RomanNum{1}, adversarial training with a gradient reversal layer reduces view identifiability in the reference representations. In Stage~\RomanNum{2}, these references are frozen to guide view-specific learning and concatenated with the resulting representations for reconstruction and clustering, with cross-view cluster alignment encouraging consistent assignments. Experiments on four benchmark datasets demonstrate ACGRL's superior clustering performance over representative methods.
Our main contributions are summarized as follows:
\begin{itemize}
    \item We introduce adversarial consistency learning with gradient reversal to reduce view identifiability in  shared representations.
    
    \item We use frozen shared representations to supervise view-specific representation learning and concatenate both components for reconstruction and clustering, with cross-view cluster alignment encouraging consistent assignments.

    \item Experiments on four benchmark datasets demonstrate that ACGRL outperforms representative multi-view clustering methods.
\end{itemize}

\section{Related work}
\label{Section_related_work}

\subsection{Multi-view Clustering}

Multi-view clustering (MVC) seeks a common cluster structure
across heterogeneous views.  Existing methods model cross-view relations
through subspace or graph learning, while deep approaches employ
reconstruction, self-expression, and representation alignment to learn
nonlinear latent structures
\cite{ICASSP_1,ICASSP_2}.
For example, EPFMVC learns view-specific representations and adaptively
fuses them according to inter-view relations \cite{EPFMVC}, whereas
self-expressive approaches learn a shared affinity structure in a unified
latent space \cite{Self_ExpressiveMetrixLearning}.
ACGRL employs adversarial training to encourage cross-view consistency in the learned representations, which are then frozen to provide fixed supervision for view-specific representation learning, with the aim of separating view-specific information from cross-view common information.
\begin{figure*}[!t]  
    \centering
    \renewcommand{\figurename}{{Figure.}}  
    \includegraphics[width=\linewidth]{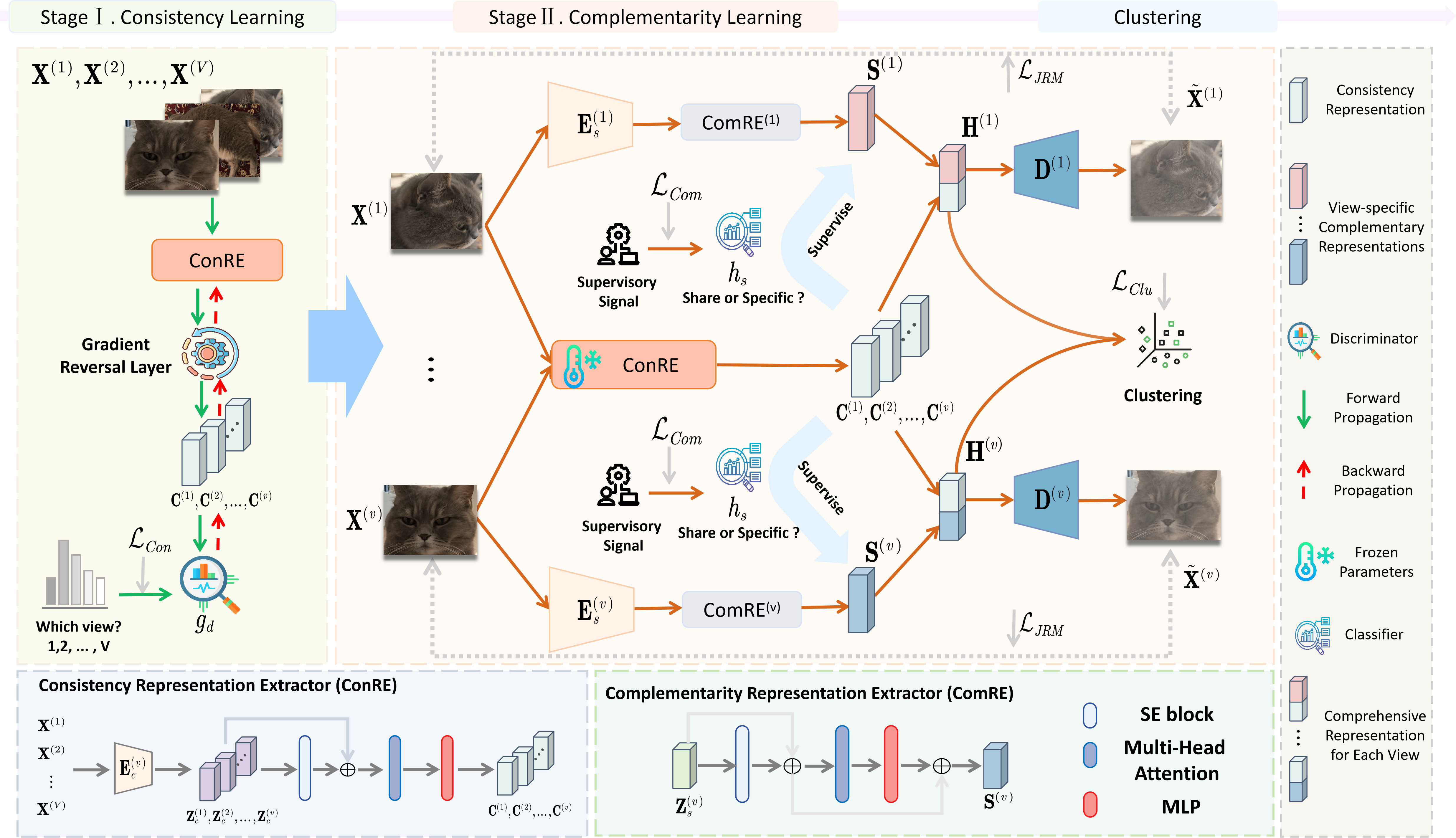} 
    \caption{Overview of ACGRL. Stage I adversarially learns view-invariant
    representations, which are frozen to provide consistency supervision for
    view-specific learning in Stage II. The resulting representations are
    integrated for reconstruction and clustering.}
    \label{ACGRL}
\end{figure*}
\subsection{Multi-view Representation Disentanglement}
Multi-view representation disentanglement aims to separate shared and view-specific information to improve clustering performance \cite{DFL_Net}. Multi-VAE disentangles shared and view-specific factors through variational modeling \cite{Multi-VAE}, while MRDD reduces their redundancy through distilled disentangling \cite{Disentanglement_Distillation}. However, separating shared and view-specific information does not necessarily eliminate view-identifying information from the shared component. ACGRL explicitly targets such information through adversarial consistency learning and uses the resulting frozen representations to guide
view-specific representation learning.

\section{Method}
\label{Section_methods}

\textbf{Notation}: We introduce Adversarial Consistency-Guided Representation Learning (ACGRL) for multi-view clustering, as illustrated in~\Cref{ACGRL}. Let $\mathcal{X}=\{\mathbf{X}^{(v)}\}_{v=1}^{V}$ denote a multi-view dataset, where $\mathbf{X}^{(v)}\in\mathbb{R}^{n\times d^{(v)}}$ is the feature matrix of the $v$-th view. ACGRL learns consistent and complementary representations for clustering.

\subsection{Adversarial Consistency Learning}
\label{sec:adversarial_consistency}

In Stage I, we learn shared representations whose source views are difficult
to identify. For each view, the consistency representation extractor maps
the input into a latent representation:
\begin{equation}
    \mathbf{C}^{(v)}
    =
    f_c^{(v)}(\mathbf{X}^{(v)}),
    \qquad v=1,\ldots,V,
    \label{eq:consistency_encoder}
\end{equation}
where $f_c^{(v)}$ denotes the Consistency Representation Extractor (ConRE), which comprises an input encoder,  an SE-based feature recalibration block, and a multi-head self-attention block with residual connections. 

To explicitly reduce view-identifiable information, we introduce a shared discriminator $g_d$ coupled with the consistency extractors
through a gradient reversal layer (GRL). The adversarial objective is
\begin{equation}
    \mathcal{L}_{\mathrm{Con}}
    =
    \frac{1}{NV}
    \sum_{v=1}^{V}\sum_{i=1}^{N}
    \operatorname{CE}
    \left(
    g_d(\mathcal{R}_{\lambda}(\mathbf{c}^{(v)}_i)),v
    \right),
    \label{eq:adversarial_consistency}
\end{equation}
where $\mathcal{R}_{\lambda}$ acts as an identity mapping during forward
propagation but reverses the gradient during backpropagation:
$\partial\mathcal{R}_{\lambda}/\partial\mathbf{x}=-\lambda\mathbf{I}$.
Consequently, the discriminator learns to identify the source view, whereas
the consistency extractors are optimized to make this identification
difficult, thereby encouraging view-invariant representations.

The reversal coefficient is progressively increased as
\begin{equation}
    \lambda(p)=\frac{2}{1+\exp(-\beta p)}-1,
    \label{eq:grl_schedule}
\end{equation}
where $p\in[0,1]$ denotes the normalized training progress and $\beta$
controls the growth rate of the adversarial signal.

\subsection{Consistency-Supervised View-Specific Learning}
\label{sec:view_specific_learning}

After Stage \RomanNum{1}, we freeze the consistency extractors to keep the learned representations \(\mathbf C^{(v)}\) fixed, allowing them to serve as anchors that supervise the subsequent learning of view-specific representations. For each view, an independently parameterized view-specific extractor maps the original input to
\begin{equation}
    \mathbf S^{(v)}
    =
    f_s^{(v)}(\mathbf X^{(v)}),
    \qquad v=1,\ldots,V,
    \label{eq:view_specific_encoder}
\end{equation}
where $f_s^{(v)}$ represents the Complementarity Representation Extractor (ComRE), which employs an independent view-specific encoder and the
same types of SE and multi-head attention blocks as $f_c$. It further
combines the recalibrated and attention-enhanced features using a learnable
weight $\alpha_v$.

To distinguish view-specific variations from the fixed shared anchors, we
introduce an auxiliary classifier $h_s$.The auxiliary classifier $h_s$ assigns the frozen shared representations to class $0$ and the view-specific representations to their respective view classes $1,\ldots,V$. Its objective is
\begin{equation}
\begin{aligned}
\mathcal{L}_{\mathrm{Com}}
=
\frac{1}{V}\sum_{v=1}^{V}
\Big[
&\epsilon\,
\operatorname{CE}\big(h_s(\mathbf S^{(v)}),v\big)
+(1-\epsilon)\,
\operatorname{CE}\big(h_s(\mathbf C^{(v)}),0\big)
\\
&+\eta\,
\mathcal H\big(h_s(\mathbf S^{(v)})\big)
\Big].
\end{aligned}
\label{eq:view_specific_loss}
\end{equation}
The fixed representations $\mathbf C^{(v)}$ provide a common reference,
while the view labels encourage $\mathbf S^{(v)}$ to retain variations
specific to each view. The entropy term further encourages confident
view assignments. Reconstruction and clustering objectives are subsequently
introduced to preserve task-relevant information in these representations.

\subsection{Reconstruction and Clustering Objectives}
\label{sec:clustering_objective}

For each view, the view-invariant and view-specific representations are
concatenated to form the comprehensive representation
$\mathbf{H}^{(v)}=[\mathbf{C}^{(v)},\mathbf{S}^{(v)}]\in\mathbb{R}^{N\times 2o}$.
To retain the information required to characterize each view, a decoder
reconstructs the original input from $\mathbf{H}^{(v)}$:
\begin{equation}
    \mathcal{L}_{\mathrm{JRM}}
    =
    \frac{1}{V}\sum_{v=1}^{V}
    \left\|
    D^{(v)}(\mathbf{H}^{(v)})-\mathbf{X}^{(v)}
    \right\|_{F}^{2}.
    \label{eq:reconstruction}
\end{equation}
A set of independently parameterized view-specific clustering heads
$\{g_p^{(v)}\}_{v=1}^{V}$ produces the soft cluster assignments:
\begin{equation}
    \mathbf{Q}^{(v)}
    =
    g_p^{(v)}(\mathbf{H}^{(v)})
    \in\mathbb{R}^{N\times K},
    \qquad v=1,\ldots,V,
    \label{eq:cluster_assignment}
\end{equation}
where the final layer of each $g_p^{(v)}$ applies a softmax operation, and
$\mathbf{q}_j^{(v)}=\mathbf{Q}_{:,j}^{(v)}\in\mathbb{R}^{N}$ denotes the
assignment vector of the $j$-th cluster across all instances in view $v$. To align the clustering structures across views, assignment vectors corresponding to the same cluster are treated as positive pairs, while those corresponding to different clusters are treated as negative pairs. For a pair of views $(v,u)$, the contrastive clustering loss is defined as
\begin{equation}
    \ell_{vc}^{(v, u)} = -\frac{1}{K} \sum_{j = 1}^K log \frac{\mathrm{e}^{sim(\textbf{Q}^{(v)}_{:, j}, \textbf{Q}^{(u)}_{:, j})/\tau}}{\sum_{m=v,u}\sum_{k=1, k\neq j}^{K} \mathrm{e}^{sim(\textbf{Q}^{(v)}_{:,j}, \textbf{Q}^{(m)}_{:, k})/\tau}},
\end{equation}

where $\operatorname{sim}(\cdot,\cdot)$ denotes cosine similarity and
$\tau$ is the temperature parameter. The loss is averaged over all ordered
view pairs:
\begin{equation}
    \mathcal{L}_{\mathrm{clu}}
    =
    \frac{1}{V(V-1)}
    \sum_{v=1}^{V}
    \sum_{\substack{u=1, u\neq v}}^{V}
    \ell_{\mathrm{vc}}^{(v,u)}.
    \label{eq:clustering_loss}
\end{equation}
Finally, the view-specific assignment matrices are uniformly averaged to
obtain the consensus cluster assignment:
\begin{equation}
    \bar{\mathbf{Q}}
    =
    \frac{1}{V}\sum_{v=1}^{V}\mathbf{Q}^{(v)}.
    \label{eq:consensus_assignment}
\end{equation}
The predicted cluster label of the $i$-th instance is then given by
\begin{equation}
    \hat{y}_i
    =
    \underset{1\leq j\leq K}{\arg\max}\,
    \bar{\mathbf{Q}}_{ij},
    \label{eq:cluster_prediction}
\end{equation}
where $\bar{\mathbf{Q}}_{i,:}\in\mathbb{R}^{K}$ denotes the consensus
assignment probabilities of the $i$-th instance over the $K$ clusters.

        
        

        





\section{Experiments}
\label{Section_Experiment}

\subsection{Datasets and Baseline}
We evaluate ACGRL on four widely used multi-view datasets. NGs \cite{NGs} contains 500 samples described by 3 views and grouped into 5 classes. BBCSport \cite{BBCSport} contains 544 samples described by 2 views and grouped into 5 classes. Cora \cite{cora} contains 2,708 samples described by 2 views and grouped into 7 classes. Hdigit \cite{Hdigit} contains 10,000 samples described by 2 views and grouped into 10 classes. 

We compare ACGRL with nine representative methods. MFLVC \cite{Unfusion_CrossView} separates reconstruction and
contrastive consistency learning into different feature levels.
DCP \cite{DCP} combines mutual-information-based contrastive
learning with cross-view prediction.
MetaViewer \cite{Metaviewer} learns unified representations
through a meta-learning-based uniform-to-specific framework.
DealMVC \cite{Linear_Fusion_Dealmvc} captures cross-view
consistency through global and local contrastive calibration.
GCFAggMVC \cite{GCFAggMVC} combines global and cross-view
feature aggregation with structure-guided contrastive learning.
DFL-Net \cite{DFL_Net} learns consistent and complementary
representations through contrastive and complementarity
regularization.
DDMVC \cite{DDMVC} combines contrastive consistency learning
with sample diversity and feature decorrelation.
GAVIM \cite{GAVIM} integrates variational information
maximization, local geometry preservation, and Gramian
alignment for imputation-free incomplete multi-view clustering.
DMVCS \cite{DMVCS} combines representation disentanglement
with semantic relevance alignment and multi-hop neighbor
contrastive learning.
All baselines are evaluated using the parameter settings
recommended in their respective papers.

\subsection{Experimental Setup}
We evaluate clustering performance using clustering accuracy (ACC), adjusted Rand index (ARI), and normalized mutual information (NMI). ACGRL is optimized using Adam with a cosine-annealing learning-rate schedule. The initial learning rate is set to $10^{-4}$, except for
the clustering module, which uses $10^{-3}$. Stages I and II are trained for 300 and 400 epochs, respectively.


\begin{table*}[!t]
\centering
\small
\caption{Clustering performance comparison on four multi-view datasets
(\%). The best results are highlighted in \textbf{bold} with a pink
background, while the second-best results are \underline{underlined}
with a gray background.}
\label{tab:clustering_comparison}

\setlength{\tabcolsep}{2.5pt}
\renewcommand{\arraystretch}{1.08}

\begin{tabular}{
c | 
ccc@{\hspace{3pt}}
ccc@{\hspace{3pt}}
ccc@{\hspace{3pt}}
ccc
}
\toprule

\multirow{2}{*}{\textbf{Method}}
& \multicolumn{3}{c}{\textbf{NGs}}
& \multicolumn{3}{c}{\textbf{Hdigit}}
& \multicolumn{3}{c}{\textbf{BBCSport}}
& \multicolumn{3}{c}{\textbf{Cora}} \\

\cmidrule(lr){2-4}
\cmidrule(lr){5-7}
\cmidrule(lr){8-10}
\cmidrule(lr){11-13}

& ACC & ARI & NMI
& ACC & ARI & NMI
& ACC & ARI & NMI
& ACC & ARI & NMI \\
\midrule

MFLVC (CVPR 2022) 
& 49.00 & 38.28 & 50.20
& 99.74
& \cellcolor{gray!20}\underline{99.42}
& 99.17
& 64.52 & 38.77 & 46.22
& 46.42 & 26.31 & 37.00 \\

DCP (TPAMI 2022) 
& 22.52 & 0.14 & 4.64
& 99.62 & 99.16 & 98.78
& 42.65 & 5.53 & 13.96
& 23.46 & -1.18 & 3.84 \\

MetaViewer (CVPR 2023) 
& 31.30 & 3.61 & 17.25
& 82.35 & 72.22 & 77.19
& 43.24 & 13.44 & 17.40
& 41.07 & 14.92 & 21.83 \\

DealMVC (ACM MM 2023)
& \cellcolor{gray!20}\underline{91.60}
& \cellcolor{gray!20}\underline{79.84}
& 78.51
& \cellcolor{gray!20}\underline{99.80}
& 99.34
& \cellcolor{gray!20}\underline{99.33}
& \cellcolor{gray!20}\underline{80.70}
& \cellcolor{gray!20}\underline{60.05}
& \cellcolor{gray!20}\underline{65.59}
& 52.70 & 27.82 & 36.96 \\

GCFAggMVC (CVPR 2023)
& 68.20 & 44.52 & 50.94
& 97.44 & 94.34 & 92.96
& 66.54 & 40.18 & 55.71
& 49.34 & 23.31 & 30.98 \\

DFL-Net (TKDE 2025) 
& 90.40 & 78.00 & 77.39
& 98.90 & 97.57 & 96.74
& 61.03 & 38.85 & 43.36
& 54.91
& \cellcolor{gray!20}\underline{31.72}
& \cellcolor{gray!20}\underline{39.74} \\

DDMVC (PR 2025)
& 90.40
& 78.47
& \cellcolor{gray!20}\underline{79.19}
& 99.43 & 98.74 & 98.19
& 69.67 & 46.46 & 53.41
& \cellcolor{gray!20}\underline{55.24}
& 30.80
& 36.20 \\

GAVIM (AAAI 2026) 
& 49.20 & 21.92 & 27.16
& 81.33 & 84.30 & 78.37
& 31.25 & 2.07 & 3.08
& 23.97 & 0.20 & 0.54 \\

DMVCS (TKDE 2026) 
& 47.20 & 13.78 & 17.37
& 89.96 & 86.79 & 91.05
& 37.68 & 2.28 & 8.08
& 30.58 & 0.32 & 1.34 \\

\midrule

\textbf{Ours}
& \cellcolor{pink!30}\textbf{97.00}
& \cellcolor{pink!30}\textbf{92.26}
& \cellcolor{pink!30}\textbf{93.31}
& \cellcolor{pink!30}\textbf{99.95}
& \cellcolor{pink!30}\textbf{99.89}
& \cellcolor{pink!30}\textbf{99.86}
& \cellcolor{pink!30}\textbf{93.58}
& \cellcolor{pink!30}\textbf{87.34}
& \cellcolor{pink!30}\textbf{85.10}
& \cellcolor{pink!30}\textbf{64.58}
& \cellcolor{pink!30}\textbf{39.58}
& \cellcolor{pink!30}\textbf{43.61} \\

\bottomrule
\end{tabular}

\end{table*}

\begin{table}[!t]
\centering
\caption{Comparison of different loss-function combinations on the
clustering task.}
\label{tab:loss_ablation}

\setlength{\tabcolsep}{1.5pt}
\renewcommand{\arraystretch}{1.08}

\resizebox{\columnwidth}{!}{%
\begin{tabular}{c|c cccc|c}
\toprule

\textbf{Dataset}

& \textbf{Metric (\%)}
& \textbf{w/o $\mathcal{L}_{\mathrm{Con}}$}
& \textbf{w/o $\mathcal{L}_{\mathrm{Com}}$}
& \textbf{w/o $\mathcal{L}_{\mathrm{JRM}}$}
& \textbf{w/o $\mathcal{L}_{\mathrm{Clu}}$}
& \textbf{Ours} \\
\midrule









\multirow{3}{*}{\textbf{BBCSport}}
& ACC
& 87.16
& 92.66
& 39.45
& 35.78
& \textbf{93.58} \\

& ARI
& 71.69
& 81.95
& 2.17
& 0.00
& \textbf{87.34} \\

& NMI
& 77.59
& 81.47
& 6.64
& 0.00
& \textbf{85.10} \\

\midrule

\multirow{3}{*}{\textbf{Cora}}
& ACC
& 59.04
& 58.30
& 31.37
& 40.22
& \textbf{64.58} \\

& ARI
& 32.49
& 32.02
& 2.31
& 11.56
& \textbf{39.58} \\

& NMI
& 39.42
& 36.61
& 4.34
& 15.36
&  \textbf{43.61} \\

\bottomrule
\end{tabular}%
}

\end{table}

\subsection{Experimental Results and Analysis}
\label{Performance}
We compare ACGRL with nine representative multi-view clustering methods as shown in~\Cref{tab:clustering_comparison}. ACGRL achieves the highest ACC on all four datasets, exceeding the strongest baseline on each dataset by 5.40, 0.15, 12.88, and 9.34 percentage points on NGs, Hdigit, BBCSport, and Cora, respectively. The improvements vary across datasets, with larger gains on NGs, BBCSport, and Cora and a marginal gain on Hdigit. 

\subsection{Ablation Study}
\label{Ablation}
\textbf{Loss Function}: As shown in \Cref{tab:loss_ablation}, removing any of the
four objectives reduces ACC, ARI, and NMI on both Cora
and BBCSport. These results support the contribution
of each objective to clustering performance within
the complete two-stage framework.

\begin{figure}[h]
    \centering
    \begin{subfigure}[b]{0.48\linewidth}
        \centering
        \includegraphics[width=\linewidth]{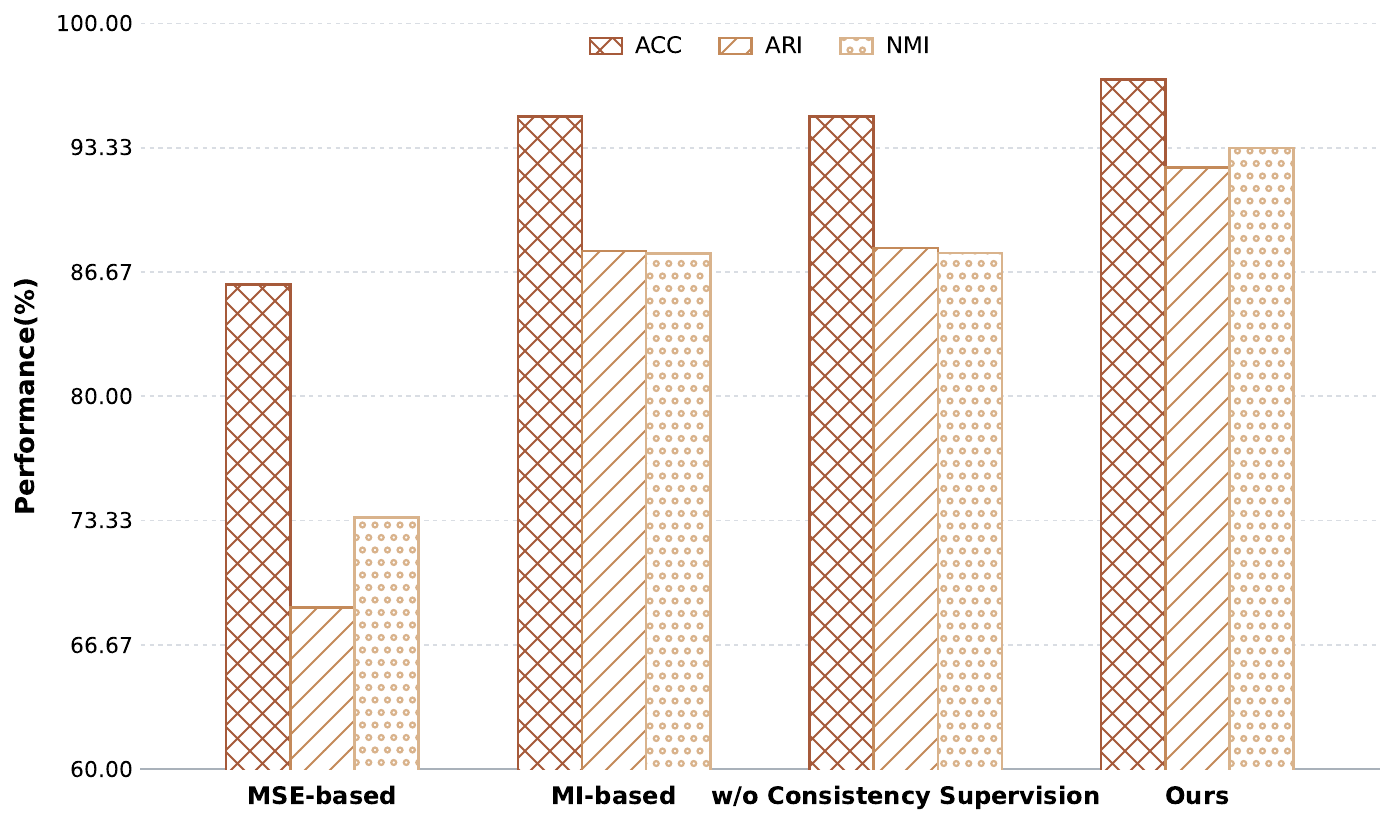}
        \caption{NGs}
        \label{fig:a}
    \end{subfigure}
    \hfill
    \begin{subfigure}[b]{0.48\linewidth}
        \centering
        \includegraphics[width=\linewidth]{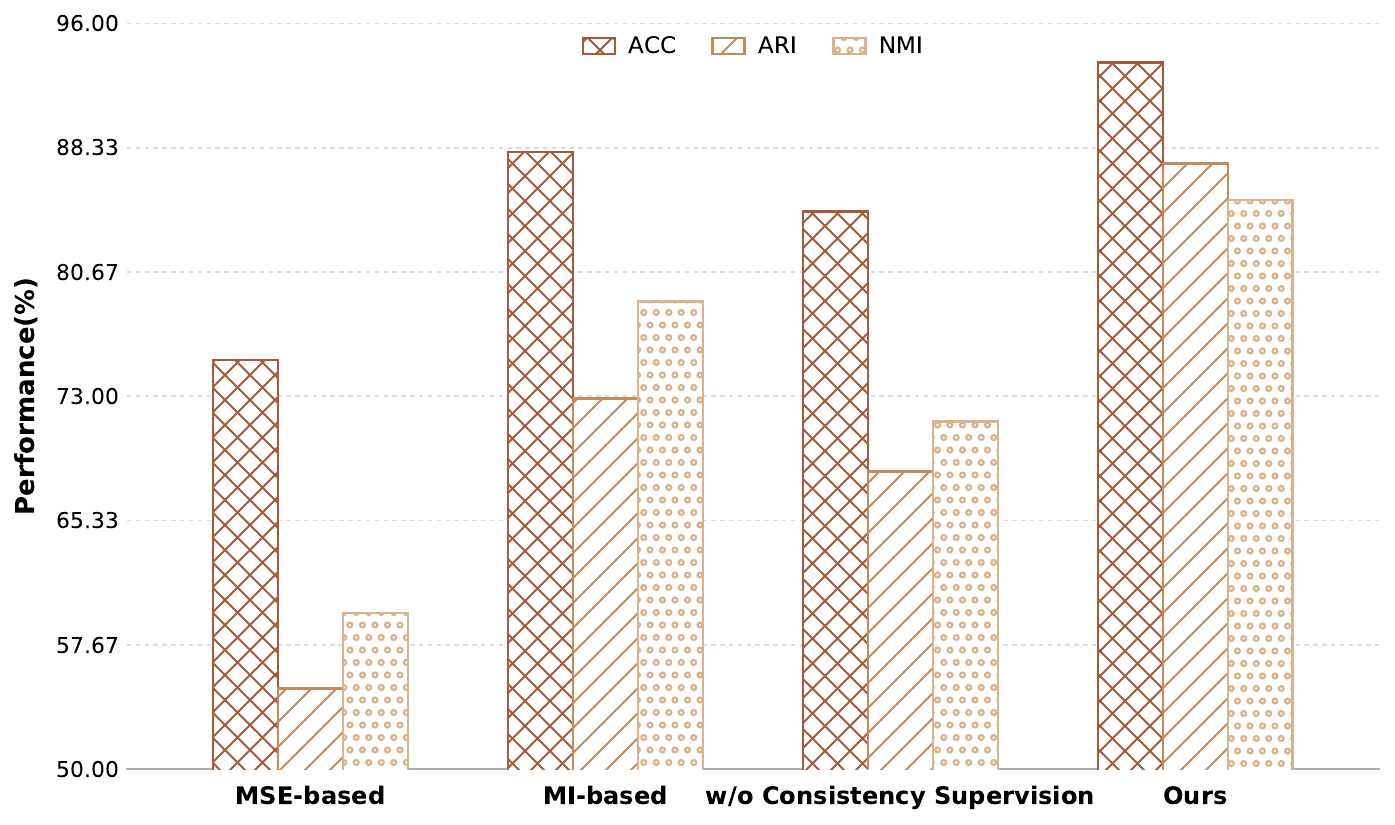}
        \caption{BBCSport}
        \label{fig:b}
    \end{subfigure}
    \caption{ Ablation study on different components.}
    \label{fig:component_ablation}
\end{figure}

\textbf{Component analysis.}  We evaluate three variants on NGs and BBCSport (\Cref{fig:component_ablation}).
The MSE- and MI-based variants replace adversarial
consistency learning in Stage I with direct representation
alignment and mutual-information maximization, respectively.
The \emph{w/o Consistency Supervision} variant removes
fixed-reference guidance in Stage II.
Full ACGRL outperforms all three variants on both datasets,
supporting the effectiveness of adversarial consistency
learning and fixed-reference guidance within our framework.

\subsection{Representation Separation Analysis}
We assess representational similarity using centered kernel
alignment (CKA) \cite{HSIC} and canonical correlation analysis
(CCA) \cite{CCA}. For a compact summary, we define the
CKA--CCA Separation Index (CCSI) as
\begin{equation}
\operatorname{CCSI}(\mathbf A,\mathbf B)
=1-\frac{
\operatorname{CKA}(\mathbf A,\mathbf B)
+\overline{\rho^2}(\mathbf A,\mathbf B)
}{2},
\end{equation}
where $\overline{\rho^2}$ denotes the mean squared canonical
correlation. Higher CCSI values indicate lower average
similarity under the two measures, rather than directly
establishing semantic disentanglement.

\begin{figure}[t]
    \centering
    \begin{subfigure}[b]{0.48\linewidth}
        \centering
        \includegraphics[width=\linewidth]{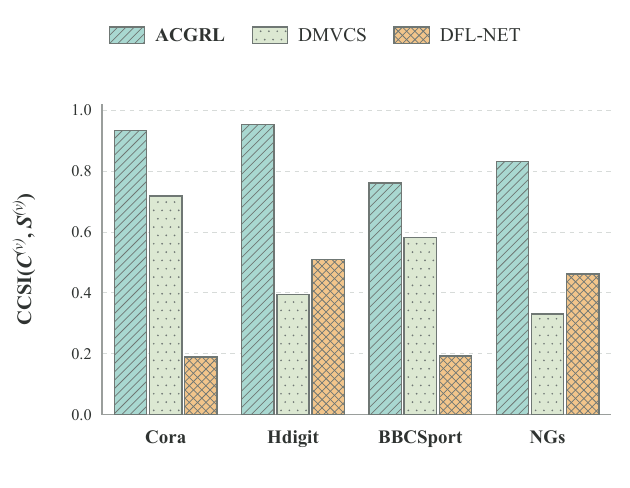}
        \caption{Intra-view consistency--specific separation.}
        \label{fig:ccsi_intra}
    \end{subfigure}
    \hfill
    \begin{subfigure}[b]{0.48\linewidth}
        \centering
        \includegraphics[width=\linewidth]{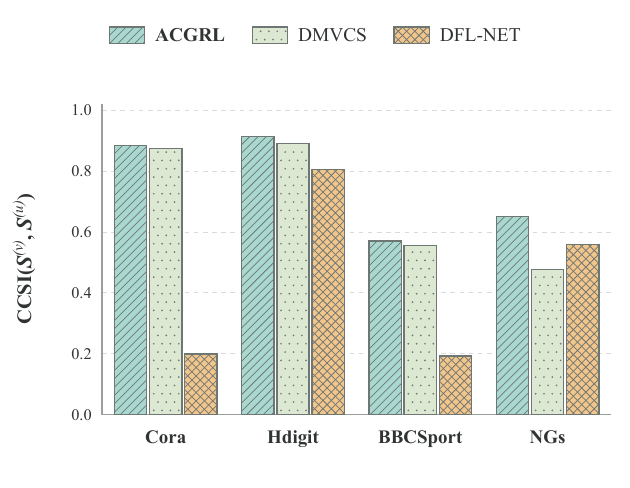}
        \caption{Cross-view specific-feature separation.}
        \label{fig:ccsi_inter}
    \end{subfigure}
    \caption{Disentanglement comparison using CCSI.}
    \label{fig:disentanglement}
\end{figure}

We compute CCSI between $\mathbf C^{(v)}$ and $\mathbf S^{(v)}$
within each view, and between $\mathbf S^{(v)}$ and
$\mathbf S^{(u)}$ across views ($v\neq u$).
As shown in \Cref{fig:disentanglement}, ACGRL achieves higher
CCSI values than DMVCS and DFL-NET in both settings across
all four datasets. These results indicate lower average
representational similarity under the selected measures,
complementing the clustering and ablation analyses.

\subsection{Visualization Analysis}
\label{Visualization}
\begin{figure}[h]
\centering

\begin{subfigure}[b]{0.48\linewidth}
\centering
\includegraphics[width=0.48\linewidth]{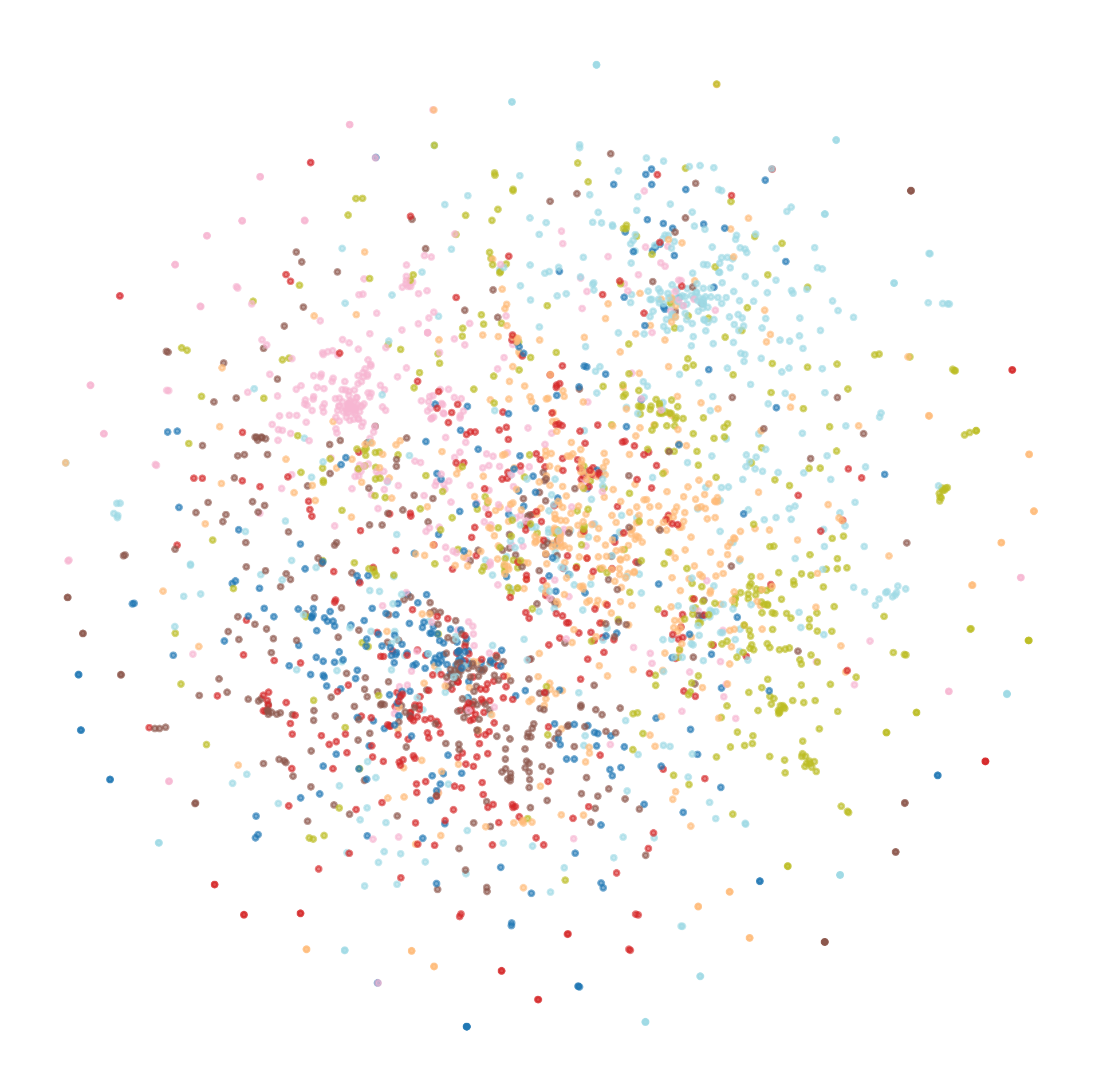}
\hfill
\includegraphics[width=0.48\linewidth]{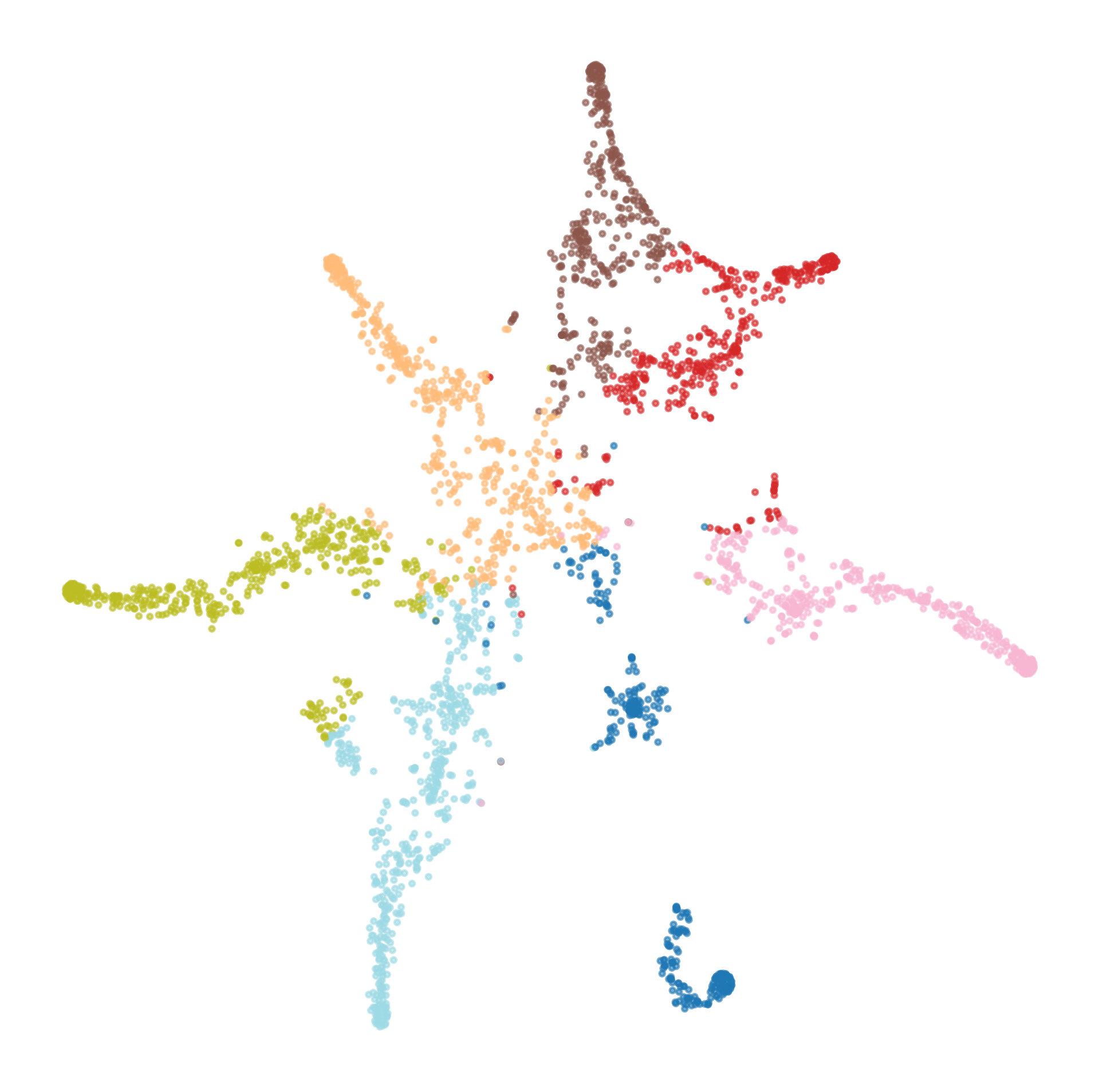}
\caption{Cora: raw (left) and learned (right) features.}
\label{fig:cora_tsne}
\end{subfigure}
\hfill
\begin{subfigure}[b]{0.48\linewidth}
\centering
\includegraphics[width=0.48\linewidth]{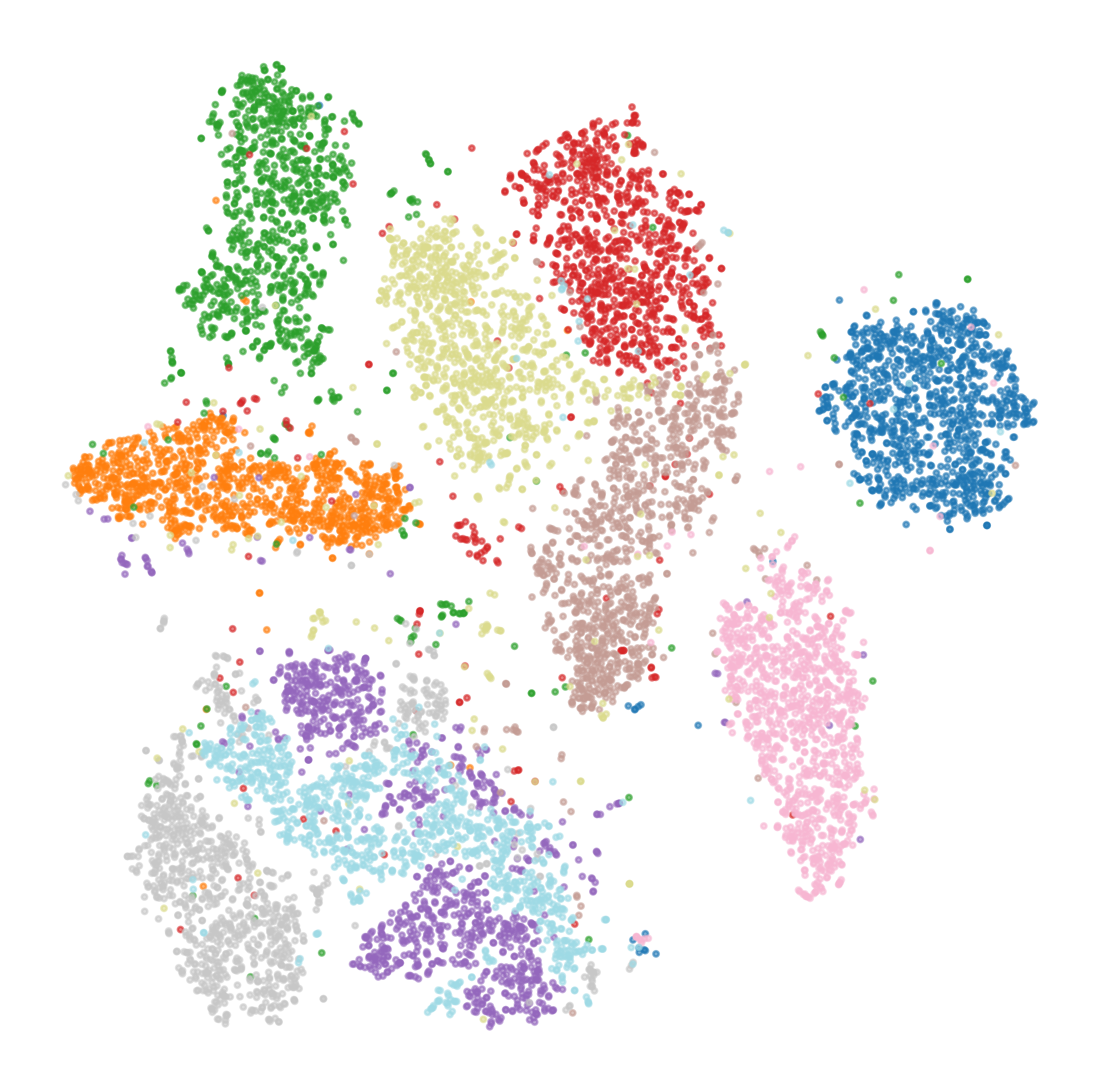}
\hfill
\includegraphics[width=0.48\linewidth]{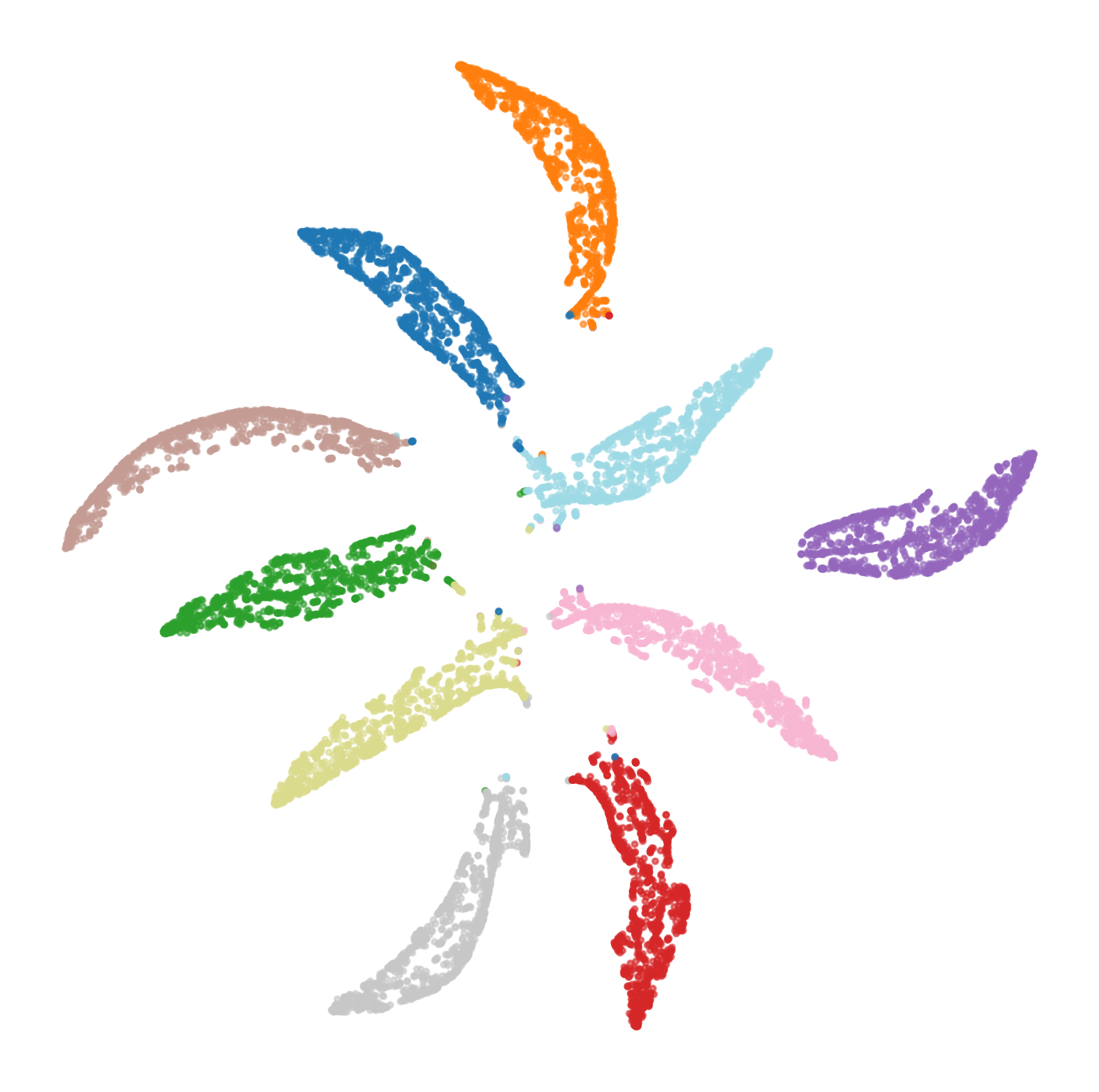}
\caption{Hdigit: raw (left) and learned (right) features.}
\label{fig:hdigit_tsne}
\end{subfigure}

\caption{t-SNE visualizations of the raw features and learned clustering representations on the Cora and Hdigit datasets.}
\label{fig:tsne}
\end{figure}

We employ t-SNE to qualitatively assess the representations learned by ACGRL. As shown in~\Cref{fig:tsne}, for both Cora and Hdigit, the left and right plots visualize the raw features and the learned representations, respectively. Compared with the highly overlapping distributions of the raw features, the learned representations exhibit clearer cluster structures, with improved inter-cluster separation on Cora and more compact and distinguishable clusters on Hdigit. These observations qualitatively demonstrate that ACGRL learns more cluster-discriminative representations.

\section{Conclusion}
\label{Section_Conclusion}
This paper presented ACGRL, a two-stage framework for disentangling consistent and complementary representations in multi-view clustering. ACGRL employs a view discriminator with a gradient reversal layer to suppress view-identifying information in shared representations. These representations
are then frozen and used as fixed guidance for separating
shared and view-specific information in the second stage.
Experiments on four benchmark datasets demonstrate that
ACGRL consistently outperforms representative baselines,
while ablation and representation separation analyses
further support the effectiveness of its core designs.

\section*{\small Compliance with Ethical Standards}
This study uses publicly available benchmark datasets and does not
involve human or animal subjects; no ethical approval was required.

\section*{\small Acknowledgments}
This work was supported by the Natural Science Foundation of Fujian Province, China, under Grants 2026J001408 and 2024J01067, and by the National Natural Science Foundation of China under Grant U1805263. The authors have no relevant financial or non-financial interests to disclose.

\bibliographystyle{IEEEbib}
\bibliography{strings,refs}

@ArtifactSoftware{R,
    title = {R: A Language and Environment for Statistical Computing},
    author = {{R Core Team}},
    organization = {R Foundation for Statistical Computing},
    address = {Vienna, Austria},
    year = {2019},
    url = {https://www.R-project.org/},
}

@inproceedings{Self_ExpressiveMetrixLearning,
  title={Learning a self-expressive network for subspace clustering},
  author={Zhang, Shangzhi and You, Chong and Vidal, Ren{\'e} and Li, Chun-Guang},
  booktitle={Proceedings of the IEEE/CVF Conference on Computer Vision and Pattern Recognition},
  pages={12393--12403},
  year={2021}
}

@article{DFL_Net,
  title={DFL-Net: Disentangled Feature Learning Network for Multi-view Clustering},
  author={Chen, Zhe and Wu, Xiao-Jun and Xu, Tianyang and Kittler, Josef},
  journal={IEEE Transactions on Knowledge and Data Engineering},
  year={2025},
  publisher={IEEE}
}

@article{DCP,
  title={Dual contrastive prediction for incomplete multi-view representation learning},
  author={Lin, Yijie and Gou, Yuanbiao and Liu, Xiaotian and Bai, Jinfeng and Lv, Jiancheng and Peng, Xi},
  journal={IEEE Transactions on Pattern Analysis and Machine Intelligence},
  volume={45},
  number={4},
  pages={4447--4461},
  year={2022},
  publisher={IEEE}
}

@inproceedings{Disentanglement_Distillation,
  title={Rethinking multi-view representation learning via distilled disentangling},
  author={Ke, Guanzhou and Wang, Bo and Wang, Xiaoli and He, Shengfeng},
  booktitle={Proceedings of the IEEE/CVF Conference on Computer Vision and Pattern Recognition},
  pages={26774--26783},
  year={2024}
}

@inproceedings{Metaviewer,
  title={Metaviewer: Towards a unified multi-view representation},
  author={Wang, Ren and Sun, Haoliang and Ma, Yuling and Xi, Xiaoming and Yin, Yilong},
  booktitle={Proceedings of the IEEE/CVF Conference on Computer Vision and Pattern Recognition},
  pages={11590--11599},
  year={2023}
}

@inproceedings{Linear_Fusion_Dealmvc,
  title={Dealmvc: Dual contrastive calibration for multi-view clustering},
  author={Yang, Xihong and Jiaqi, Jin and Wang, Siwei and Liang, Ke and Liu, Yue and Wen, Yi and Liu, Suyuan and Zhou, Sihang and Liu, Xinwang and Zhu, En},
  booktitle={Proceedings of the 31st ACM international conference on multimedia},
  pages={337--346},
  year={2023}
}

@article{DDMVC,
  title={Deep multi-view clustering with diverse and discriminative feature learning},
  author={Xu, Junpeng and Meng, Min and Liu, Jigang and Wu, Jigang},
  journal={Pattern Recognition},
  volume={161},
  pages={111322},
  year={2025},
  publisher={Elsevier}
}

@inproceedings{Unfusion_CrossView,
  title={Multi-level feature learning for contrastive multi-view clustering},
  author={Xu, Jie and Tang, Huayi and Ren, Yazhou and Peng, Liang and Zhu, Xiaofeng and He, Lifang},
  booktitle={Proceedings of the IEEE/CVF conference on computer vision and pattern recognition},
  pages={16051--16060},
  year={2022}
}

@inproceedings{EPFMVC,
  title={Enhanced then Progressive Fusion with View Graph for Multi-View Clustering},
  author={Dong, Zhibin and Liu, Meng and Wang, Siwei and Liang, Ke and Zhang, Yi and Liu, Suyuan and Jin, Jiaqi and Liu, Xinwang and Zhu, En},
  booktitle={Proceedings of the Computer Vision and Pattern Recognition Conference},
  pages={15518--15527},
  year={2025}
}

@inproceedings{Multi-VAE,
  title={Multi-VAE: Learning disentangled view-common and view-peculiar visual representations for multi-view clustering},
  author={Xu, Jie and Ren, Yazhou and Tang, Huayi and Pu, Xiaorong and Zhu, Xiaofeng and Zeng, Ming and He, Lifang},
  booktitle={Proceedings of the IEEE/CVF international conference on computer vision},
  pages={9234--9243},
  year={2021}
}

@inproceedings{cora,
  title={Co-clustering of multi-view datasets: a parallelizable approach},
  author={Bisson, Gilles and Grimal, Cl{\'e}ment},
  booktitle={2012 IEEE 12th international conference on data mining},
  pages={828--833},
  year={2012},
  organization={IEEE}
}

@inproceedings{NGs,
  title={An improved co-similarity measure for document clustering},
  author={Hussain, Syed Fawad and Bisson, Gilles and Grimal, Cl{\'e}ment},
  booktitle={2010 ninth international conference on machine learning and applications},
  pages={190--197},
  year={2010},
  organization={IEEE}
}

@article{DMVCS,
  title={Disentangled Contrastive Multi-view Clustering via Semantic Relevance Invariance},
  author={Li, Pengyuan and Chang, Dongxia and Wang, Yiming and Kong, Zisen and Kong, Linhua and Zhao, Yao},
  journal={IEEE Transactions on Knowledge and Data Engineering},
  year={2026},
  publisher={IEEE}
}

@inproceedings{GAVIM,
  title={Geometry-Aware Variational Information Maximization for Deep Incomplete Multi-view Clustering},
  author={Chen, Wenlan and Gao, Lu and Wang, Daoyuan and Guo, Fei and Liang, Cheng},
  booktitle={Proceedings of the AAAI Conference on Artificial Intelligence},
  volume={40},
  number={24},
  pages={20289--20297},
  year={2026}
}

@inproceedings{GCFAggMVC,
  title={Gcfagg: Global and cross-view feature aggregation for multi-view clustering},
  author={Yan, Weiqing and Zhang, Yuanyang and Lv, Chenlei and Tang, Chang and Yue, Guanghui and Liao, Liang and Lin, Weisi},
  booktitle={Proceedings of the IEEE/CVF conference on computer vision and pattern recognition},
  pages={19863--19872},
  year={2023}
}

@inproceedings{HSIC,
  title={Measuring statistical dependence with Hilbert-Schmidt norms},
  author={Gretton, Arthur and Bousquet, Olivier and Smola, Alex and Sch{\"o}lkopf, Bernhard},
  booktitle={International conference on algorithmic learning theory},
  pages={63--77},
  year={2005},
  organization={Springer}
}

@incollection{CCA,
  title={Relations between two sets of variates},
  author={Hotelling, Harold},
  booktitle={Breakthroughs in statistics: methodology and distribution},
  pages={162--190},
  year={1992},
  publisher={Springer}
}

@inproceedings{ICASSP_1,
  title={Graph-Guided Contrastive Learning for Incomplete Multi-View Clustering with Consistent Global Graph},
  author={Wang, Zhepeng and Zhang, Zhenghao and Zong, Tianyu and Chen, Feng and Xie, Jun},
  booktitle={ICASSP 2026-2026 IEEE International Conference on Acoustics, Speech and Signal Processing (ICASSP)},
  pages={2581--2585},
  year={2026},
  organization={IEEE}
}

@inproceedings{ICASSP_2,
  author       = {Nan Li and
                  Songlin Du},
  title        = {Underlying-Complementarity and Surrounding-Correspondence for Multi-View
                  Clustering},
  booktitle    = {{IEEE} International Conference on Acoustics, Speech and Signal Processing,
                  {ICASSP} 2024, Seoul, Republic of Korea, April 14-19, 2024},
  pages        = {7975--7979},
  publisher    = {{IEEE}},
  year         = {2024},
  url          = {https://doi.org/10.1109/ICASSP48485.2024.10448475},
  doi          = {10.1109/ICASSP48485.2024.10448475},
  bibsource    = {dblp computer science bibliography, https://dblp.org}
}

@article{Hdigit,
  title={Dual contrast-driven deep multi-view clustering},
  author={Cui, Jinrong and Li, Yuting and Huang, Han and Wen, Jie},
  journal={IEEE Transactions on Image Processing},
  volume={33},
  pages={4753--4764},
  year={2024},
  publisher={IEEE}
}

@article{BBCSport,
  author       = {Dino Ienco and
                  C{\'{e}}line Robardet and
                  Ruggero G. Pensa and
                  Rosa Meo},
  title        = {Parameter-less co-clustering for star-structured heterogeneous data},
  journal      = {Data Min. Knowl. Discov.},
  volume       = {26},
  number       = {2},
  pages        = {217--254},
  year         = {2013},
  url          = {https://doi.org/10.1007/s10618-012-0248-z},
  doi          = {10.1007/S10618-012-0248-Z},
  bibsource    = {dblp computer science bibliography, https://dblp.org}
}

\end{document}